%% file: main.tex
\pdfoutput=1

\documentclass[11pt]{article}
\usepackage{arabtex}
\def\endabstract{\egroup}
\usepackage[]{acl2023}

\usepackage{times}
\usepackage{latexsym}
\definecolor{codegreen}{rgb}{0,0.6,0}
\definecolor{codegray}{rgb}{0.5,0.5,0.5}
\definecolor{codepurple}{rgb}{0.58,0,0.82}
\definecolor{backcolour}{rgb}{0.95,0.95,0.92}
\usepackage{listings}
\usepackage{xcolor}
\lstdefinestyle{mystyle}{
    backgroundcolor=\color{backcolour},   
    commentstyle=\color{codegreen},
    keywordstyle=\color{magenta},
    numberstyle=\tiny\color{codegray},
    stringstyle=\color{codepurple},
    basicstyle=\ttfamily\footnotesize,
    breakatwhitespace=false,         
    breaklines=true,                 
    captionpos=b,                    
    keepspaces=true,                 
    numbers=left,                    
    numbersep=5pt,                  
    showspaces=false,                
    showstringspaces=false,
    showtabs=false,                  
    tabsize=2
}

\usepackage[T1]{fontenc}

\usepackage{graphicx}
\usepackage{microtype}
\usepackage{booktabs}
\usepackage{subcaption}
\usepackage{multirow}
\usepackage{array}
\usepackage{makecell}
\usepackage{natbib}
\usepackage{titlepic}
\usepackage{xspace}
\usepackage{utf8}
\setcode{utf8}
\usepackage{inconsolata}

\newcommand{\icon}[1]{\includegraphics[height=1.5cm]{#1}}
\newcommand{\taqyim}{\texttt{Taqyim}\xspace}

\title{
\protect\icon{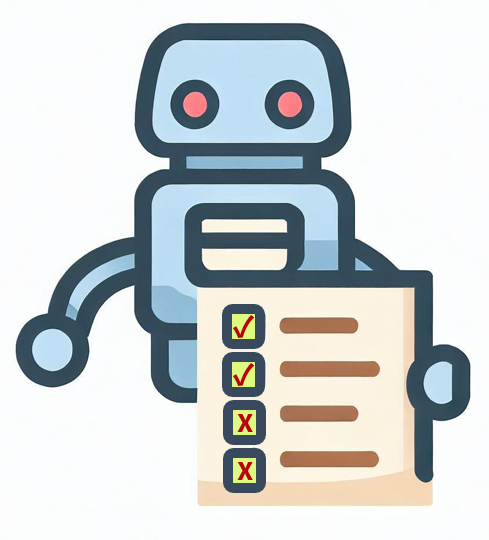}

\textbf{Taqyim: Evaluating Arabic NLP Tasks Using ChatGPT Models}}

\author{
\textbf{Zaid Alyafeai}\thanks{$\;\;$Equal Contribution}$\;^{,1,\gamma}$~~~ 
\textbf{Maged S. Alshaibani}\footnotemark[1]$\;^{,1}$~~~ 
\textbf{Badr AlKhamissi}\footnotemark[1]$\;^{,2}$~~~ \\
\textbf{Hamzah Luqman}$^{1,3}$~~~ 
\textbf{Ebrahim Alareqi}$^4$~~~
\textbf{Ali Fadel}$^2$~~~
\\
\\
$^1$ King Fahd University of Petroleum and Minerals,  Saudi Arabia \\
$^2$ ARBML \\
$^3$ SDAIA-KFUPM Joint Research Center for Artificial Intelligence
\\
$^4$ Volvo Cars R\&D Tech Center, United States \\
\\
$^{\gamma}$  Corresponding Author: \nolinkurl{g201080740@kfupm.edu.sa}
}

\begin{document}
\maketitle
\begin{abstract}
Large language models (LLMs) have demonstrated impressive performance on various downstream tasks without requiring fine-tuning, including ChatGPT, a chat-based model built on top of LLMs such as GPT-3.5 and GPT-4. Despite having a lower training proportion compared to English, these models also exhibit remarkable capabilities in other languages. In this study, we assess the performance of GPT-3.5 and GPT-4 models on seven distinct Arabic NLP tasks: sentiment analysis, translation, transliteration, paraphrasing, part of speech tagging, summarization, and diacritization. Our findings reveal that GPT-4 outperforms GPT-3.5 on five out of the seven tasks. Furthermore, we conduct an extensive analysis of the sentiment analysis task, providing insights into how LLMs achieve exceptional results on a challenging dialectal dataset. Additionally, we introduce a new Python interface\footnote{\url{https://github.com/ARBML/Taqyim}} that facilitates the evaluation of these tasks effortlessly. 
\end{abstract}

\section{Introduction}

The emergence of foundation models \cite{Bommasani2021OnTO} in recent years has instigated a transformative shift within the field of Natural Language Processing (NLP). The conventional practice of pre-training and subsequently fine-tuning a model specifically for a given task has been shown to no longer be necessary on some tasks. Instead, research has shown that a sufficiently large model trained on vast amounts of data is capable of achieving comparable, and sometimes better, performance compared to task-specific models. 
However, despite its success across numerous NLP tasks, fine-tuned models remain superior in a variety of domains. For instance, its proficiency in solving elementary mathematical operations has been found to be lacking \cite{davis2023mathematics, frieder2023mathematical, gilson2022well}, while its performance in tasks involving commonsense reasoning has demonstrated much room for improvement \cite{davis2023mathematics, guo2023close}. Moreover, concerns have been raised regarding the language coverage of this purportedly general-purpose language model \cite{bang2023multitask, jiao2023chatgpt, lu2022trip} leading to worse performance on non-English languages.

In this paper, we present a comprehensive evaluation of the performance of two ChatGPT-based models, namely GPT-3.5 and GPT-4, across seven crucial Arabic NLP tasks. The evaluation is conducted with the aim of assessing the capabilities of these emerging foundation models and comparing their performance against state-of-the-art (SoTA) counterparts. The selected tasks for this study encompass a diverse range of NLP applications, including summarization, diacritization, part of speech tagging, sentiment analysis, transliteration, machine translation, and paraphrasing. The findings of our investigation reveal a notable disparity between the performance of the ChatGPT models and that of their Arabic-specific counterparts across most tasks, with the exception of summarization, where both ChatGPT models exhibit superior performance compared to existing SoTA approaches. Furthermore, the evaluation results indicate that GPT-3.5 outperforms GPT-4 on two out of the seven tasks, specifically summarization and diacritization. 

Furthermore, to gain deeper insights into the performance of the ChatGPT-based models, we conduct a comprehensive case study focusing on the Sentiment Analysis task. This investigation encompasses an examination of the impact of various factors, including temperature tuning, prompt engineering, and the effect of different numbers of few-shot demonstrations within the context, on the overall performance of the task, in addition to closely analyzing the outputs generated by the models. Moreover, in the context of the diacritization task, we provide fine-grained results across seven distinct domains, allowing for a more granular evaluation of the models' diacritization capabilities. This overall analysis sheds light on the current state of Arabic NLP in relation to the foundation models such as ChatGPT, highlighting both the potential and the existing gaps that warrant further exploration and improvement in Arabic NLP. 

In conclusion, we introduce a novel Python library, named \taqyim, derived from the Arabic word for "evaluation." This library is designed to enhance the evaluation process and is developed as an extension of the OpenAI \texttt{evals} library, incorporating three fundamental principles: (1) Ease of Use, (2) Robustness, and (3) Debugging. By prioritizing user-friendly functionalities, \taqyim aims to streamline the evaluation workflow, facilitating seamless integration and efficient utilization of the library. \taqyim is released as an open-source library, allowing the wider community to benefit from its capabilities and contribute to its ongoing development and improvement.  


\begin{figure*}[ht!]
    \centering
    \includegraphics[width=0.8\textwidth]{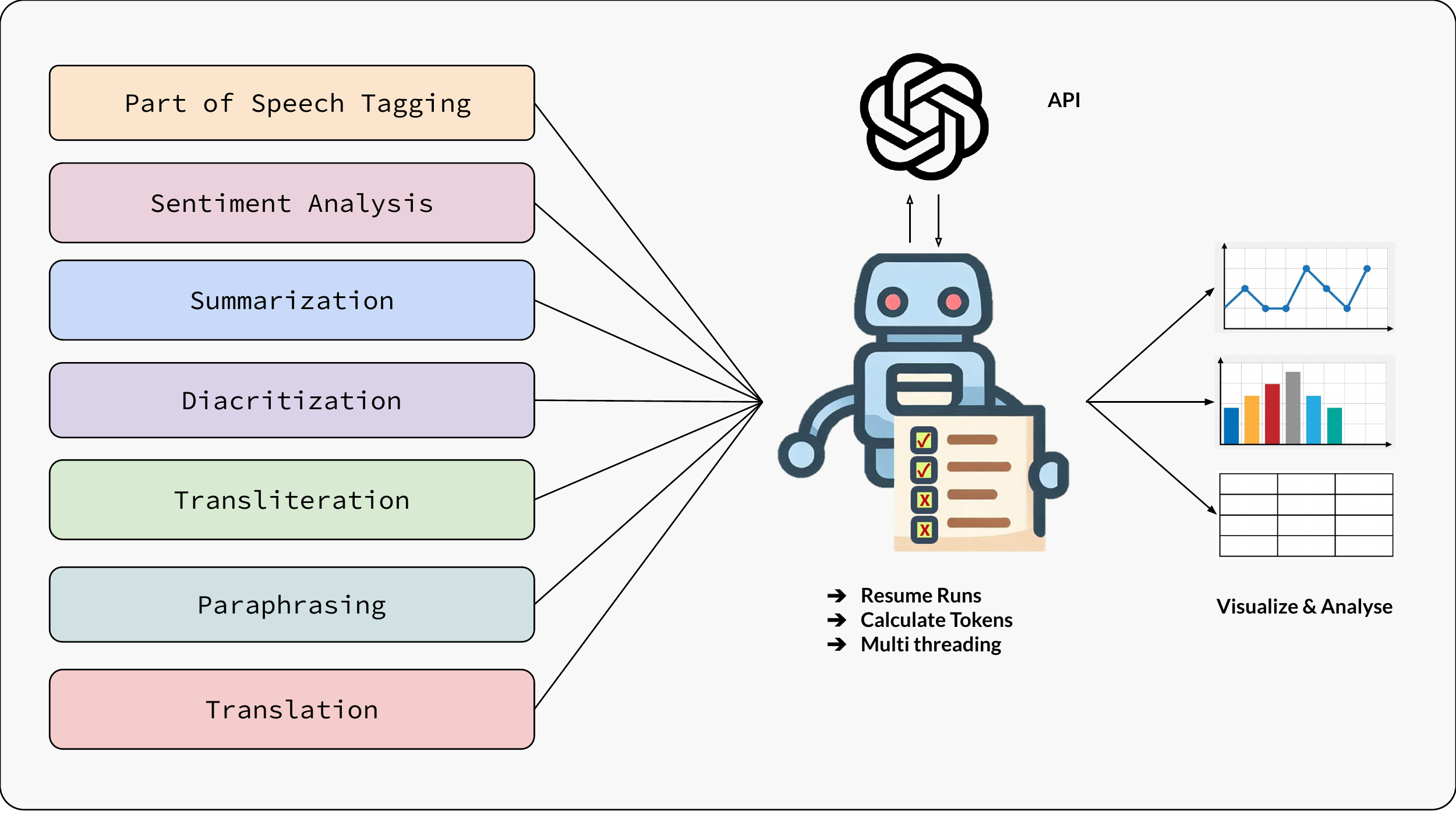}
    \caption{\taqyim Pipeline.}
    \label{fig:taqyim_pipeline}
\end{figure*}


\section{Related Work}

\paragraph{Large Language Models.} 

Several language models have been proposed recently. One of the earliest pre-trained language models is ELMo which was proposed to model the word context \cite{peters-etal-2018-deep}. ELMo learns the word context by pre-training a two-layer bidirectional LSTM network on large data and fine-tuning it on downstream tasks. BERT followed this learning strategy with a Transformer model pre-trained on large datasets \cite{Devlin2019BERTPO}. The performance of BERT outperformed other models on several downstream tasks. This learning paradigm motivated researchers to propose either new architectures (e.g., BART \cite{lewis2019bart} and GPT-2 \cite{radford2019language}) or enhanced pre-training techniques \cite{liu2019roberta, sanh2021multitask, wang2022language}. Scaling language models in terms of model size or data used for model pre-training has shown its effectiveness in several downstream tasks \cite{zhao2023survey}. This led to the introduction of the “large language models (LLM)” term. These models are trained on large datasets and usually have billions of parameters. Such LLMs showed a better performance compared with the smaller models with similar architectures and pre-training tasks (e.g., GPT-3 \cite{brown2020language} vs GPT-2).  Recently, a significant number of LLMs have been introduced, such as GPT-3, LLaMA \cite{touvron2023llama}, PaLM \cite{chowdhery2022palm}, BLOOM \cite{muennighoff2022crosslingual}, and Chinchilla \cite{hoffmann2022training}. ChatGPT\footnote{ \url{https://openai.com/blog/chatgpt}} is one of these LLMs that was developed based on the GPT model series (GPT-3.5 and GPT-4) and showed a powerful performance with dialogue tasks. 

\paragraph{ChatGPT Evaluation} 
Following the introduction of ChatGPT, numerous studies have emerged assessing its performance across diverse tasks, encompassing machine translation \cite{jiao2023chatgpt, hendy2023good}, reasoning \cite{bang2023multitask, qin2023chatgpt}, the health care domain \cite{SLAPETA2023314, medicalchatgpt2023}, among others \cite{liu2023summary}. In one investigation, \citet{bang2023multitask} comprehensively assessed the performance of ChatGPT on eight distinct NLP tasks, employing a diverse set of 23 datasets. These tasks primarily revolved around the English language, with the exception of machine translation. The findings of this study unveiled that ChatGPT outperformed several state-of-the-art models across various NLP tasks. However, certain limitations were observed in specific scenarios, such as summarization, machine translation, particularly for low-resource languages, and reasoning capabilities.

Conversely, \citet{qin2023chatgpt} conducted an investigation revealing ChatGPT's robust performance in numerous NLP tasks, particularly highlighting its proficiency in reasoning tasks. However, the model exhibited limitations in certain tasks, such as sequence tagging. Furthermore, \citet{chan2023chatgpt} evaluated ChatGPT's capabilities in capturing inter-sentential relations, including discourse, causal, and temporal relationships. The reported outcomes demonstrated ChatGPT's adeptness in identifying and reasoning causal relationships. Conversely, the model's performance proved to be sub-optimal in tasks related to dialogue discourse parsing and detecting temporal relationships between two events. Notably, ChatGPT exhibited satisfactory detection of explicit discourse relations, yet encountered difficulties in handling implicit discourse relations.

\paragraph{Concurrent Work}
During the course of our study, two papers have been published that evaluate ChatGPT models across multiple Arabic NLP tasks. Namely, \citet{tawkat2023gptaraeval} evaluated GPT-3.5 on a variety of Arabic NLU \cite{Elmadany2022ORCAAC} and NLG tasks, and compared it against the multilingual BLOOMZ ($7.1$B) model \cite{muennighoff2022crosslingual} and the much smaller monolingual AraT5 \cite{elmadany2022arat5} fine-tuned on each respective task. Their evaluation encompassed varying numbers of few-shot demonstrations within the context, with $\textsc{n-shot}=\{0, 3, 5, 10\}$. Their results demonstrated that while GPT-3.5 exhibited superior performance compared to BLOOMZ on Arabic tasks, it still significantly trailed behind the smaller-scale, Arabic-specific finetuned model, AraT5. In another related study, \citet{abdelali2023benchmarking} conducted an evaluation of GPT-3.5 on a range of Arabic NLP and Speech processing tasks. However, their investigation lacked explicit disclosure to some of their evaluation methodology (e.g. on the diacritization task), hindering the reproducibility of their findings. In contrast, our research provides comprehensive and transparent documentation of all pertinent details. Additionally, we extend the evaluation to include GPT-4 and provide additional analysis of the models' outputs in Section \ref{sec:analysis}.


\begin{figure*}[ht!]
    \centering
    \includegraphics[width=1\linewidth]{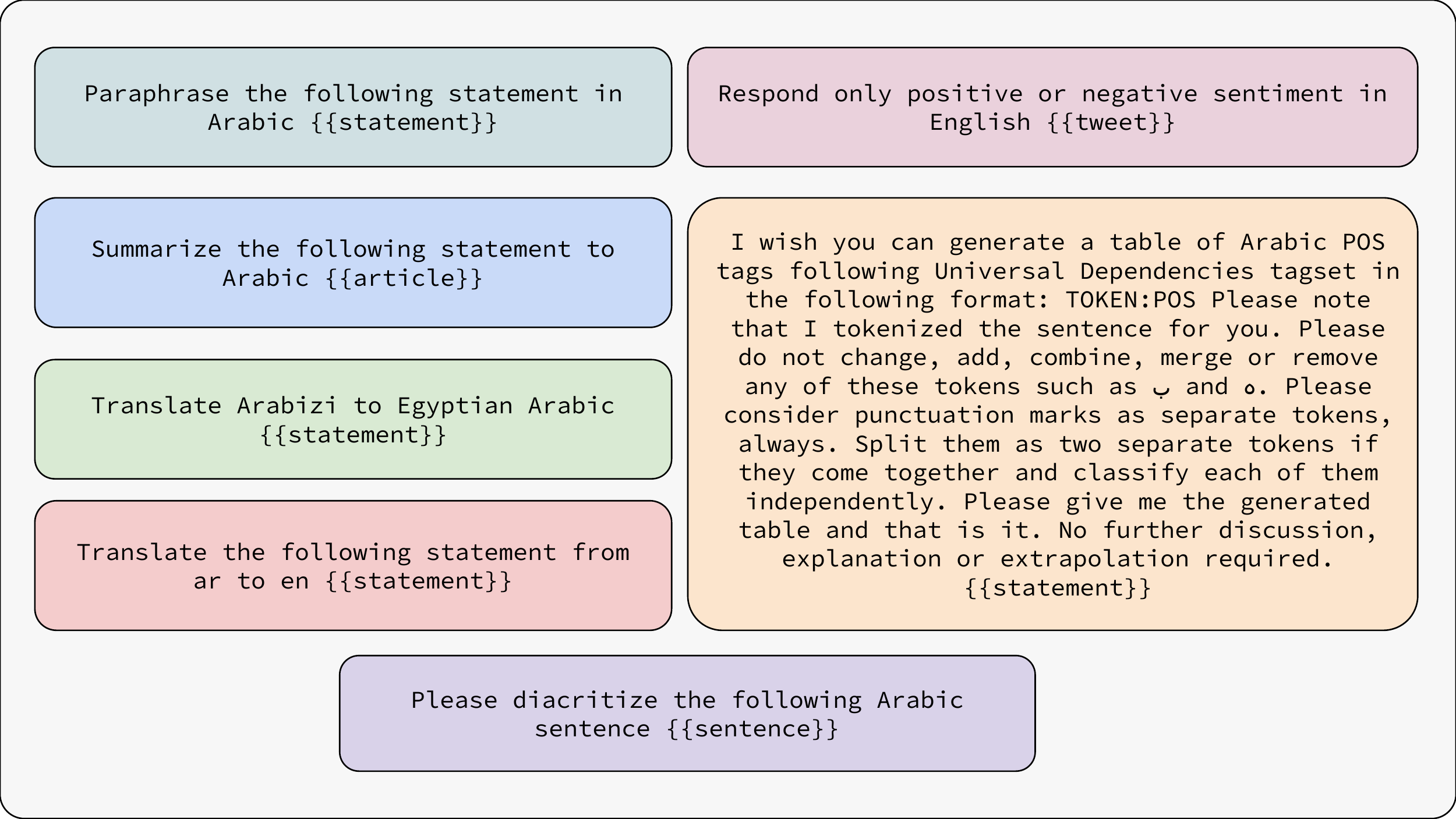}
    \caption{Prompts used for each task. The double curly braces \texttt{\{\{\}\}} indicate placeholders that are taken from the dataset to apply the prompt on.}
    \label{fig:prompts}
\end{figure*}

\section{Pipeline}

In Figure \ref{fig:taqyim_pipeline}, we highlight the pipeline for \taqyim. Given a set of tasks, we pass that to a Python interface that contacts the OpenAI API back and forth and gets the evaluation results. 
Our Python interface is built on top of a forked version of OpenAI's \texttt{evals}\footnote{\url{https://github.com/openai/evals}} library. It has four main advantages over the \texttt{evals} library, in the following paragraphs we illustrate each feature. 

\paragraph{Ease of use} The \texttt{evals} library does not have a Python interface, which makes evaluations a bit more complex. To tackle this problem, we created a Python interface, that could be used to run the \texttt{evals} codebase. In addition to that, we use the \texttt{datasets}\footnote{\url{https://github.com/huggingface/datasets}} library to provide a single hub for loading and downloading any dataset to run evaluation on by just providing the name of the dataset. The following code snippet gives an example of how to run an evaluation on a sentiment analysis dataset (AJGT).  
\begin{lstlisting}[language=Octave]
import taqyim as tq

# create the pipeline
pipeline = tq.Pipeline(
    eval_name="ajgt-test",
    dataset_name="arbml/ajgt_ubc_split",
    task_class="classification",
    task_description= "Sentiment Analysis",
    input_column_name="content",
    target_column_name="label",
    prompt="Predict the sentiment",
    api_key="<openai-key>",
    train_split="train",
    test_split="test",
    model_name="gpt-3.5-turbo-0301",
    max_samples=1,)

# run the evaluation
pipeline.run()
\end{lstlisting}

\paragraph{Robustness} The OpenAI's API counts the number of tokens as the total of the tokens required to compute the input and completion. It will return an error if the input size is greater than the model context size. To calculate that efficiently, we use the \texttt{tiktoken}\footnote{\url{https://github.com/openai/tiktoken}} library to calculate the number of tokens of the input which is subtracted from the model max context size. 
In addition to that, to make our library more robust, we allow resuming a run after an error due to stopping execution. The \texttt{resume\_from\_record} flag is used to resume any given run which hugely reduces cost. 

\paragraph{Debugging} The \texttt{evals} library shuffles the samples before sending and fetching the API results. This approach makes debugging difficult because the `sample\_id` key in the results is not in sync with the row number in the original test dataset. We force the library to send the requests sequentially which makes it easy to debug and visualize the results in a sequential manner. 

\paragraph{Analysis}
Given the output from \taqyim, we can represent our output as a data frame that can be easily used for analyzing and visualizing the output to get some useful insights about the ChatGPT completions. 

\section{Tasks}

In this section, we illustrate the datasets used for evaluating the ChatGPT models. The prompts used in evaluation are summarized in Figure \ref{fig:prompts}. We also summarize the datasets used in Table \ref{tab:datasets}.

\input{tables/datasets}

\subsection{Summarization}
In this task, we aim at predicting the summary of a given article. The summary could vary in length depending on the article. In a prompted fashion, given an \texttt{article}, we want to prompt ChatGPT to predict that \texttt{summary}. 

\paragraph{EASC} For this task, we use Essex Arabic Summary Corpus (EASC). The dataset contains 153 Arabic articles with their associated summaries \cite{el2010using}. For the sake of the evaluation, we use the \texttt{RougeL} score which calculates the similarity between the true summary and the predicted summary. EASC has been used a lot in the literature to evaluate LLMs. \cite{elmadany2022arat5} trained T5-based models (AraT5) on multiple tasks including summarization. They reported the results on EASC after fine-tuning on the train split of WikiLingua \cite{ladhak2020wikilingua}. Similarly, AT5B \cite{ghaddar2022revisiting} released T5-style models (AT5S and AT5B) that outperform AraT5 by pretraining on more cleaned data. AraMUS \cite{alghamdi2023aramus} an 11B parameter language model achieves better results compared to the previous results on EASC and gets a result of  13.3 using the RougeL score.

\paragraph{Preprocessing} The dataset contains some long articles which might exceed the model context size an hence can't be consumed as an API request. To avoid that, we truncate all the articles to a max size of 4,290 characters before sending the request.

\subsection{Diacritization}


The automatic restoration of diacritics to Arabic text is arguably one of the most important NLP tasks for the Arabic language. Diacritics play a crucial role in determining accurate word pronunciation and meaning, as they indicate vowel sounds and grammatical information. However, diacritics are often omitted in various written Arabic texts, such as social media posts, news articles, and formal documents, due to reasons like typing convenience, space limitations, or lack of standardization. Consequently, the task of Arabic diacritization aims to address this issue by automatically adding the missing diacritics to the text. This process facilitates precise interpretation and analysis of Arabic data, supporting a wide range of NLP applications, including machine translation, text-to-speech systems, and named entity recognition, among others \cite{zitouni09}.

\paragraph{WikiNews} 
We leverage the WikiNews test set introduced by \cite{darwish17} to evaluate the ChatGPT models on Arabic diacritization. It comprises 70 WikiNews articles written in Modern Standard Arabic (MSA), primarily sourced from 2013 and 2014, encompassing 7 domains: politics, economics, health, science and technology, sports, arts, and culture. To ensure a balanced representation, an equal number of articles (i.e., 10 articles) were allocated to each domain. In total, the WikiNews test set encompasses approximately $18,300$ words, providing a substantial corpus for assessing the performance of ChatGPT models across a diverse range of domains. The SoTA model on this test set leverages a sequence-to-sequence transformer-based model that is fine-tuned on a large diacritized corpus of 4.5 million tokens and employs an overlapping sliding window and a voting mechanism during inference (see next paragraph) to predict the final diacritic \cite{mubarak19-highly}. 

\paragraph{Evaluation Setup} 
Following the approach introduced by \cite{mubarak19-highly}, we adopt the overlapping context window methodology coupled with a voting mechanism to facilitate diacritic prediction for each character, as opposed to the naive approach of passing the entire sentence to the model at once. This technique involves dividing a given sentence into multiple overlapping segments, each individually presented to the model for inference. This approach proves effective, as local context often provides sufficient information for accurate inference. Consequently, identical character sequences may appear in different contexts (i.e., various segments within a single sentence), potentially resulting in different diacritized forms. To determine the definitive diacritic, we employ a popularity voting mechanism, and in cases where a tie occurs, we randomly select one of the outputs. Our implementation employs a sliding window of 20 words with a stride of 2, similar to \cite{alkhamissi-etal-2020-dd}.

\paragraph{Post Processing} Since ChatGPT models are not constrained during the generation process, they have the potential to produce invalid outputs, which may involve the addition or omission of characters or words in the generated text. Therefore, in order to effectively evaluate the model's performance on generation tasks like diacritization, we employ the following heuristic approach. For each word in the input sentence, we verify if it is present in the generated output. If the word is found, we incorporate the corresponding generated diacritics for that word. Conversely, if the word is not found, we include it in the output without diacritics. This methodology ensures that the output sentence maintains the same content as the input sentence while incorporating the appropriate diacritics.



\subsection{Part of Speech Tagging}
The part of speech tagging (POS) task is responsible for predicting the part of speech tags for a given list of tokens/words. 

\paragraph{PADT} For this task, we use \texttt{ar\_padt} split that is offered by \texttt{universal\_dependencies} and created by \cite{unv_2_7}. The subset contains $6,080$ samples for training, $909$ for validation, and $680$ for testing. The dataset contains 17 tags that can be used in multilingual settings. 
For prompting purposes, we feed the model by joining the \texttt{tokens} using space and predicting the tags in the following format \texttt{token:tag} separated by the new line character \texttt{\textbackslash n}. Encoder-based language models like BERT seem to achieve decent results for part of speech tasks \cite{kondratyuk201975}. They used a multilingual BERT language model that was pre-trained in 104 languages. Then they fine-tuned it on 75 languages by concatenating all of them together with simple softmax classifiers for the POS task. 

\paragraph{Post Processing}
We tested with a lot of prompts for GPT-3.5 model before getting descent completions that followed our required output format. The POS task is unique in this regard because we constrain the output to be in the format \texttt{token:tag}. For a given output completion, we first match the output tokens against the gold tokens and then extract their associated tags. We remove extra spaces or quotations that might result in some wrong evaluations. 

\subsection{Sentiment Analysis}
In this task, the model is prompted to predict the \texttt{label} given the \texttt{text}. We consider this task as a binary classification task where the model is supposed to predict only one of two classes which are \texttt{positive} or \texttt{negative}.

\paragraph{AJGT} We use the Arabic Jordanian General Tweets (AJGT) Corpus which consists of $1,800$ tweets from the Jordanian dialect \cite{alomari2017arabic}. Since the dataset doesn't have train and test splits, we use the splits suggested by \cite{elmadany2022arat5} which consists of $1,440$ samples for training and $360$ samples for testing. On this task, masked language models seem to achieve much better results, especially after fine-tuning on the train split of the dataset. More specifically the MARBERT \cite{abdul2020arbert} model which was pretrained on a social media dataset achieved a score of $96.11$ on the test split. 

\subsection{Transliteration}
Transliteration is the process of converting text from one writing system to another while maintaining the phonetic value of the original text. It enables approximating the pronunciation of words or names in a different writing system, allowing users to understand and vocalize them more easily. It allows non-Arabic speakers to approximate the pronunciation of Arabic words and names by using familiar Latin characters. 

\paragraph{BOLT} We used BOLT Egyptian Arabic Treebank dataset with a test set of size $6,653$. \cite{shazal2020unified} achieved the best results on the test set of that dataset with a score of $65.88$ on the BLEU score metric as reported by \cite{elmadany2022arat5}.

\subsection{Machine Translation}
In this task, the model is prompted to predict a given translation using the \texttt{source} and \texttt{target} languages, for example, from \texttt{ar} to \texttt{en}. In this task, we aim to translate from English to Arabic as a sample study. 

\paragraph{UNv1} We use the united nations version 1 (UNv1) \cite{ziemski2016united} with its test split that contains around $4,000$ Arabic-English pairs. On this test split \cite{elmadany2022arat5} achieves the best results on the BLEU metric with a score of $53.29$. 

\subsection{Paraphrasing}
Paraphrasing in NLP refers to the process of rephrasing or restating a given text or sentence while preserving its original meaning. In a prompted fashion, given a prompt in a specific language, we want to predict the paraphrased version in the same language.

\paragraph{APB} We used the Arabic Paraphrasing Benchmark (APB) with a test set of $1,010$ sentences. As a generative task, AraT5 achieves the best result with a score of $17.52$ on the blind test split.

\input{tables/results}

\section{Results}
\label{sec:results}
In this section, we go over the results of the evaluated datasets in a comprehensive fashion. In all the experiments, we use \texttt{gpt-3.5-turbo-0301} and \texttt{gpt-4-0314} versions for GPT-3.5 and GPT-4 respectively.

\subsection{Zero-shot Results}
In Table \ref{tab:results}, we summarize the results for all the tasks used in the study by applying the prompts in Figure \ref{fig:prompts}. It is also worth mentioning that those prompts are designed to serve the best results for GPT-3.5 model then used for evaluating GPT-4.0. We perform zero-shot evaluation where the model is only assumed to predict the \texttt{input} given \texttt{prompt}+\texttt{input}. 
 For each task, we show the dataset used, the test size, the metric, and a comparison between the ChatGPT models and the SoTA results. Our results show that GPT-4 outperforms GPT-3.5 in all the tasks except the summarization and diacritization tasks. Regarding summarization, the EASC dataset contains large summaries, while GPT-4 predicts conscience and compact summaries. We calculated the average length of summaries generated from GPT-3.5 and we got 429 compared to 348 characters from GPT-4. We discuss diacritization in more detail in Section \ref{subsec:diac}. For the other tasks, GPT-4 achieves the largest improvement margin over GPT-3.5 in the POS task because the model can predict the tokens in a more natural manner compared to GPT-3.5. We had to do a lot of prompt engineering to force GPT-3.5 to predict the tokens and tags in our constrained format. Regardless, both models still lack behind compared to fine-tuned models, especially for complex tasks like paraphrasing which contains a lot of dialectal examples. 

\subsection{Fine-grained Diacritization Results}
\label{subsec:diac}
Tables \ref{tab:finegrained-diac-gpt-3.5} and \ref{tab:finegrained-diac-gpt-4} present the performance outcomes of GPT-3.5 and GPT-4, respectively, on the WikiNews diacritization benchmark in relation to each domain. The obtained results indicate that the \textit{culture} domain exhibits the most favorable performance with the lowest error rate, whereas the \textit{arts} domain demonstrates the least satisfactory performance across both models. For a further breakdown of the diacritization results, we refer the reader to Appendix \ref{app:diac-results}.

It is worth noting that the ChatGPT models occasionally fail to generate diacritics for all characters in the input (as exemplified by the first word in Table \ref{tab:examples_gpt_3}), resulting in a significant increase in the error rate, particularly for Word Error Rate (WER). To improve the performance of diacritization, future research should consider incorporating multiple instructions and sampling multiple outputs for each input, followed by aggregating the results through a majority voting scheme. This approach is expected to enhance the accuracy of the models on this task and represents a promising direction for further investigation.

\input{tables/diacritization_finegrained}

\section{Sentiment Analysis: Case Study}
\label{sec:analysis}

In this section, we conduct a case study on the AJGT dataset, where we analyze it from different perspectives. While this approach can be implemented for all the tasks and models, we only use it for the classification task due to budget constraints.

\begin{figure}[!htp]
    \centering
    \includegraphics[width=0.5\textwidth]{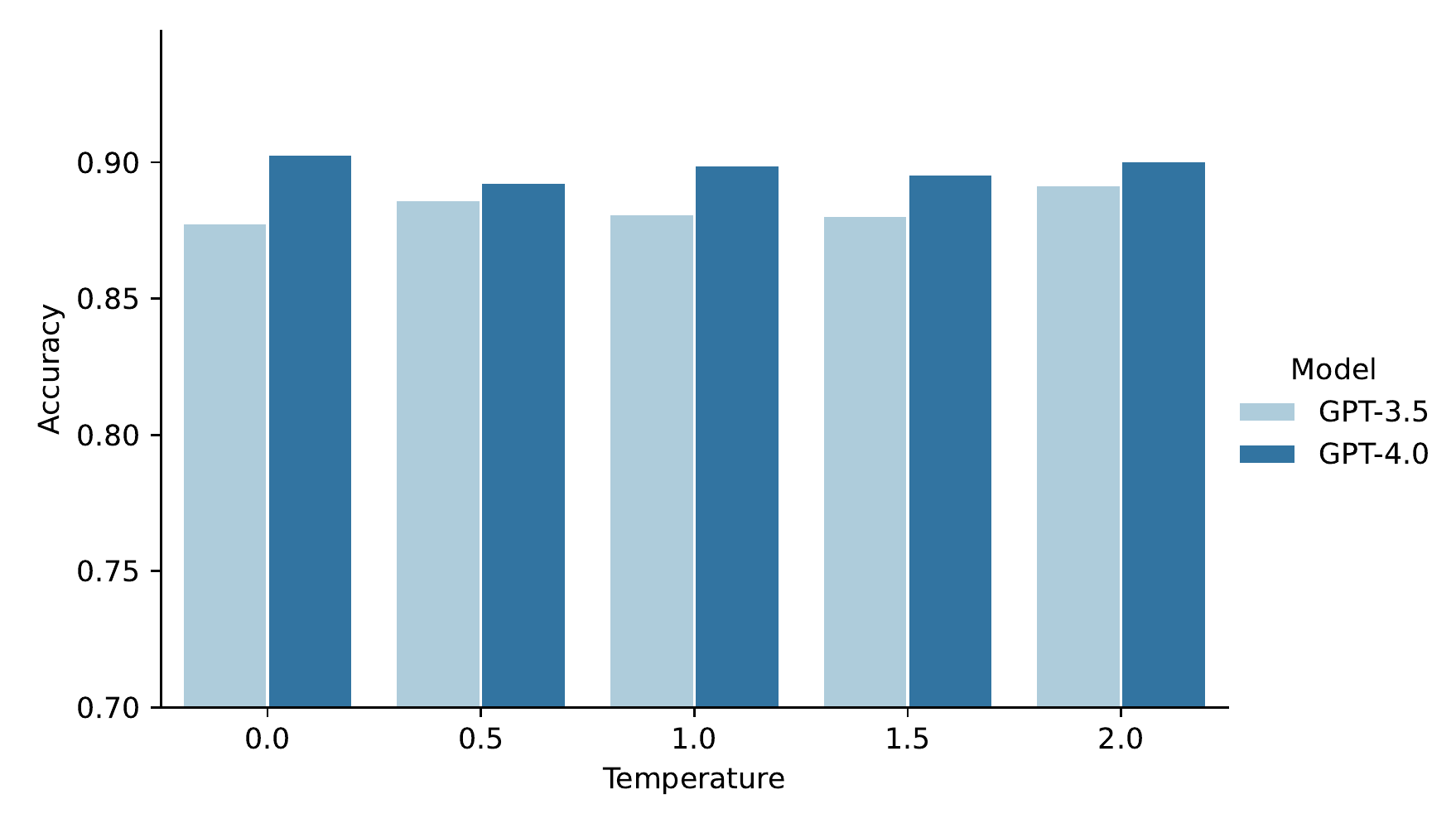}
    \caption{Temperature tuning results for the test set of AJGT. We vary the temperature from 0.0 to 2.0 and evaluate on the test set of AJGT.}
    \label{fig:temp_results}
\end{figure}

\subsection{Temperature Tuning}
Temperature is a hyperparameter used to control the creativity of a language model. Technically, it normalizes the probabilities of the softmax layer giving a chance to lower probability tokens to be selected while generating output. In both GPT models, 0 temperature is an aggressive value that takes into account only the highest probable token while a value of 2 gives the highest chance for less probable tokens to be selected while generating the output. In Figure \ref{fig:temp_results}, we show the results for different temperatures. For all the different values GPT-4 achieves better results compared to GPT-3.5. Further, zero temperature shows a noticeable performance gap between GPT-3.5 and GPT-4.0 compared to other values.

\subsection{Few-shot results}
\begin{figure}[!htp]
    \centering
    \includegraphics[width=0.5\textwidth]{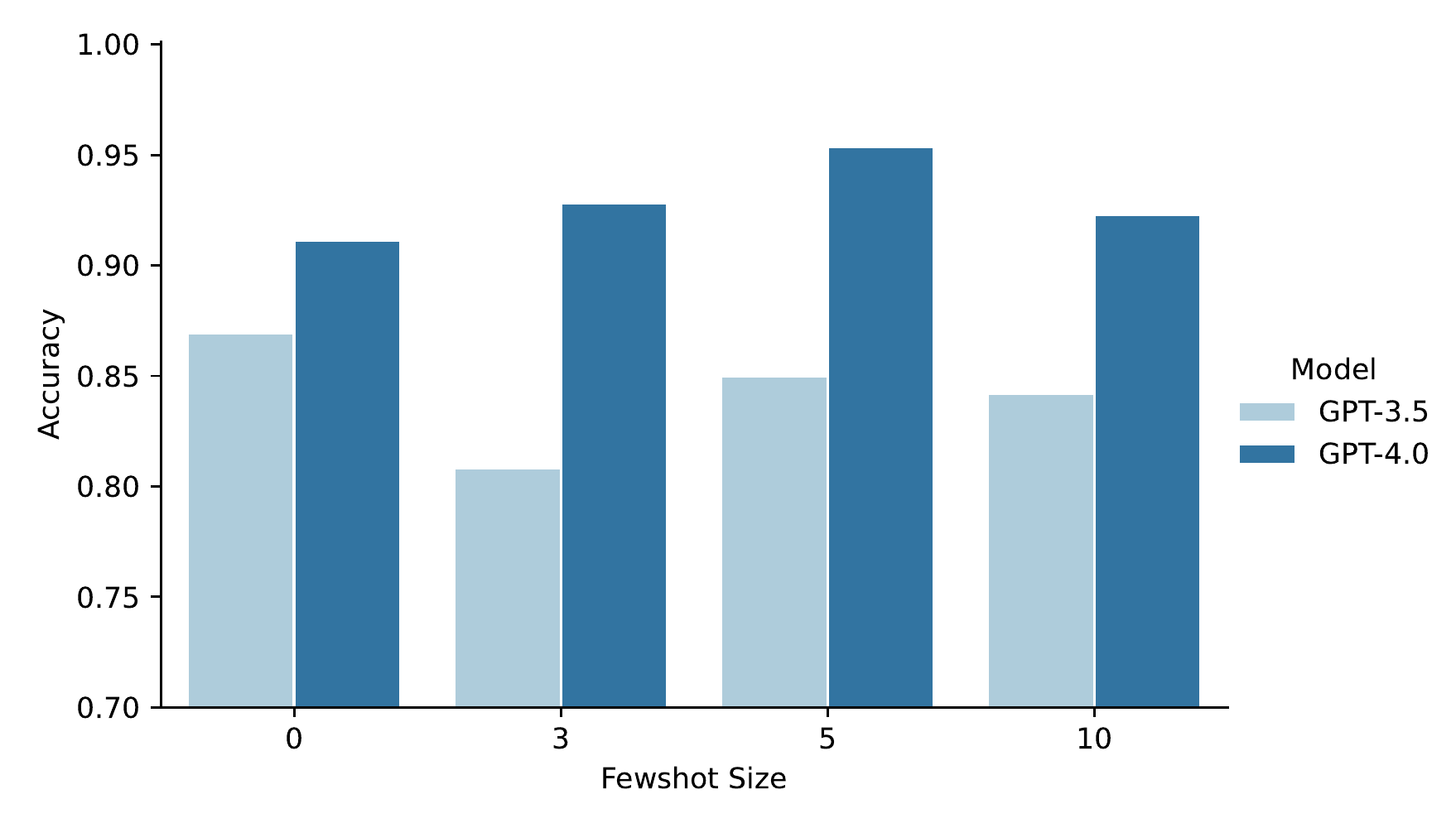}
    \caption{Fewshot results on the test set of AJGT. We evaluate both GPT-4 and GPT-3.5 using different numbers of a few shot samples [0, 3, 5, 10].}
    \label{fig:fewshot_results}
\end{figure}
Fewshot prompting defines the approach of prompting the model with multiple examples from the training corpus. In this subsection, we study the effect of few-shot size on both models. We set the temperature to be 1 as a middle-ground between creativity and aggressiveness. Figure \ref{fig:fewshot_results} shows the results of different few-shot examples applied to GPT-3.5 and GPT-4 models.
As can be seen from the Figure, increasing the few-shot examples degraded the results for GPT-3.5 while improving the results for GPT-4. We analyzed the model outputs for GPT-3.5 and found out that this model refused to give predictions on many samples for various reasons discussed in more detail in Section \ref{subsec:responses_analysis}. In contrast, more few-shot examples improved the results for GPT-4 allowing the model to reach close to the SoTA results with 5 examples. However, adding more few-shots may degrade the model's performance as in the case with 10 few-shot examples.

\input{tables/prompts}

\subsection{Prompt Engineering}
\begin{figure}[!htp]
\centering
\includegraphics[width=0.25\textwidth]{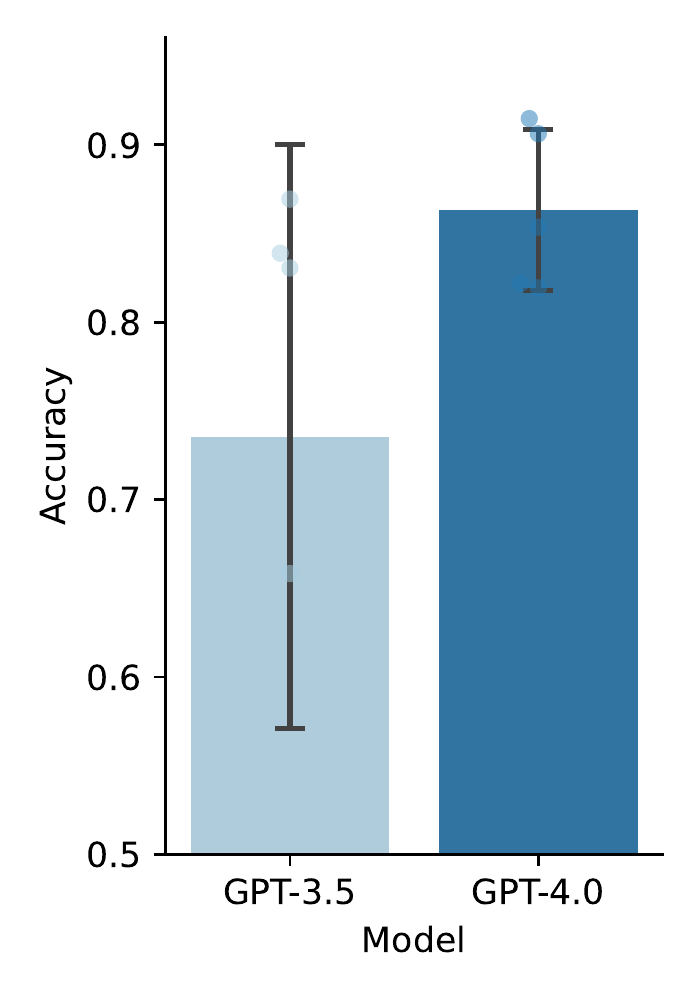}
\caption{Prompt engineering results for AJGT. For each model, we run both models on five different prompts.}
\label{fig:prompt_eng}
\end{figure}

In Figure \ref{fig:prompt_eng}, we show the results of the evaluation for multiple prompts shown in Table \ref{tab:prompts_eng}. We observe that in general, GPT-3.5 achieves a wider error range, compared to a shorter range for GPT-4. Hence, we can predict that GPT-4 is more robust against different prompts compared to GPT-3.5. 

\subsection{Responses Analysis}
\label{subsec:responses_analysis}
This section provides further details on GPT-3.5 and GPT-4.0 responses on the classification task. We studied the confusion matrices for both models and noticed that GPT-3.5 only classifies 342, 304, 317, and 312 samples for 0, 3, 5, and 10 few-shot examples respectively out of the 360 samples. We also found out that, in some cases, it may respond with a different template than the one instructed to respond with. For example, it responds with "Negative sentiment" instead of "Negative". We asked it only to respond either by "Positive" or "Negative". We went over the unique responses of GPT-3.5 to study the responses that do not contain either "Positive" or "Negative" tokens and found the following:

\begin{itemize}
    \item It cannot understand either part or whole of the sentence in Arabic. It asked the user to translate the sample to English or provide more context.
    \item It cannot determine the sentiment of the sentence by asking the user to rephrase, provide more context, or provide the text in English.
    \item Instead of providing a classification, it provides an Arabic explanation. This happens more when adding more few-shot examples.
    \item It did not respond because it thinks the provided sample is written in an inappropriate language, i.e. it is an offensive sample. However, sometimes it provides a classification with further clarification indicating that the sample is written in inappropriate or offensive language.
\end{itemize}

On the other hand, this behavior was not noticed for GPT-4. In fact, GPT-4 classifies all samples with the given two classes, either "Positive" or "Negative" in all few-shot settings except for 3 few-shot samples where it generated a new label "Neutral".
\begin{figure}[!htp]
    \centering
    \includegraphics[width=.45\textwidth]{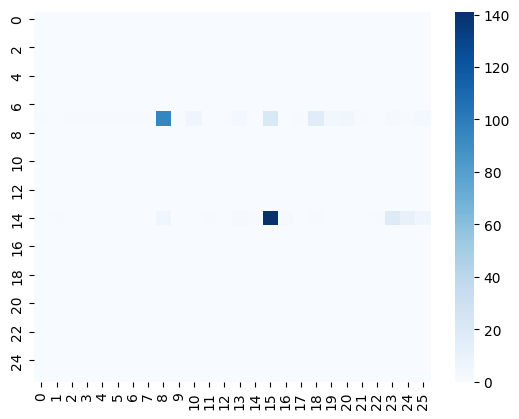}
    \caption{GPT-3.5 zero-shot responses to AJGT as a confusion matrix. Although there are only two classes in this dataset, "Positive" and "Negative", GPT3.5 responds with many variations. Numbers are removed as most of them, except the highlighted areas, are zeros. Also, responses are removed for clarity purposes.}
    \label{fig:gpt35cm}
\end{figure}

\begin{figure}
    \centering
    \includegraphics[width=.35\textwidth]{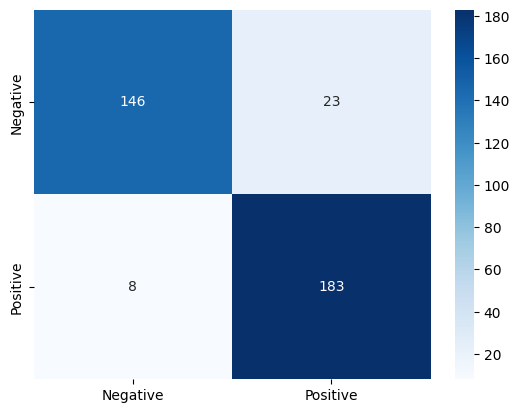}
    \caption{GPT-4 zero-shot responses to AJGT as a confusion matrix.}
    \label{fig:gpt4cm}
\end{figure}

As a result of the above analysis, GPT-3.5 has lower comprehension abilities for Arabic text compared to GPT-4. It is, also, more vulnerable to offensive samples than GPT-4. Figures \ref{fig:gpt35cm} and \ref{fig:gpt4cm} show the confusion matrix in zero-shot settings for GPT-3.5 and GPT-4, respectively. Responses counts are removed from GPT-3.5 confusion matrix as most of them are just zeros. As can be seen in the confusion matrix, the number of unique responses for GPT-3.5 is more than 20. This number further explodes when adding more few-shot examples. This can be seen in the confusion matrices of the other few-shot examples of GPT-3.5 in the Appendix. GPT-4 confusion matrices for the other few-shot examples are also added there.

\section{Conclusion}
In summary, this paper evaluated the performance of large language models (LLMs), specifically GPT-3.5 and GPT-4, on seven different Arabic NLP tasks which are sentiment analysis, translation, transliteration, paraphrasing, part of speech tagging, summarization, and diacritization. The study highlighted the impressive abilities of chat-based models like ChatGPT, which are built on LLMs, in performing these tasks in a zero-shot setting. The results demonstrated that GPT-4 outperformed GPT-3.5 on five out of the seven evaluated tasks, indicating continuous advancements in LLM technology. Additionally, we developed a new Python interface publicly available which facilitates the evaluation of these tasks with ease.

One significant aspect of this research is the exploration of sentiment analysis in a comprehensive manner. We provide valuable insights into how LLMs achieve remarkable results in dialectal tasks in a zero-shot fashion, shedding light on the underlying mechanisms of these models. Overall, this paper contributes to the growing body of knowledge regarding the capabilities of LLMs, specifically in the context of Arabic language processing. The findings not only showcase the superior performance of GPT-4 compared to its predecessor but also provide a useful tool for future evaluations in Arabic natural language processing tasks. This work paves the way for further advancements in language models and their applications across diverse languages and tasks.

\section*{Limitations}

\paragraph{Model Selection Bias} This study focuses exclusively on the evaluation of ChatGPT-based models, reflecting their increasing popularity and relevance in the field. However, it is important to acknowledge the presence of other language models (LLMs) that warrant exploration, particularly those explicitly designed with multilinguality in mind, such as BLOOMZ \cite{muennighoff2022crosslingual}. Including a wider range of LLMs in future research would provide a more comprehensive evaluation and facilitate a more informed comparison.

\paragraph{Limited Exploration of Fewshot Demonstrations} Moreover, while this work briefly explores the inclusion of few-shot demonstrations in one task, the main emphasis remains on the zero-shot scenario. Given the contextual learning capabilities of LLMs, it is reasonable to expect that incorporating few-shot demonstrations could potentially enhance model performance. However, a deeper investigation into the impact of few-shot demonstrations across multiple tasks is warranted, as this aspect remains an avenue for future research and calls for more extensive analysis.

\section*{Acknowledgement}
The authors would like to acknowledge the support received from the Saudi Data and AI Authority (SDAIA) and King Fahd University of Petroleum and Minerals (KFUPM) under the
SDAIA-KFUPM Joint Research Center for Artificial Intelligence Grant JRC-AI-RFP-05. We would like to also thank Maqsam for providing the compute to run some of our experiments.
 
\bibliographystyle{apalike}
\bibliography{references}

\newpage

\appendix

\section{Extra Responses Analysis}
\label{sec:extra_responses_analysis}

\begin{figure}
    \centering
    \includegraphics[width=.45\textwidth]{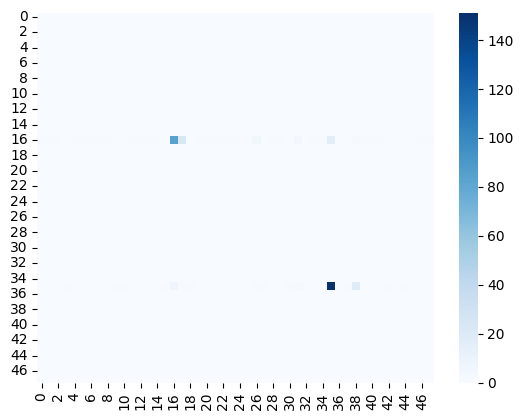}
    \caption{GPT-3.5 responses to AJGT dataset confusion matrix with 3 few-shot examples. Numbers are removed as most of them, except the highlighted areas, are zeros. Also, responses are removed for clarity purposes}
    \label{fig:gpt35cm_fewshots_3}
\end{figure}

\begin{figure}
    \centering
    \includegraphics[width=.45\textwidth]{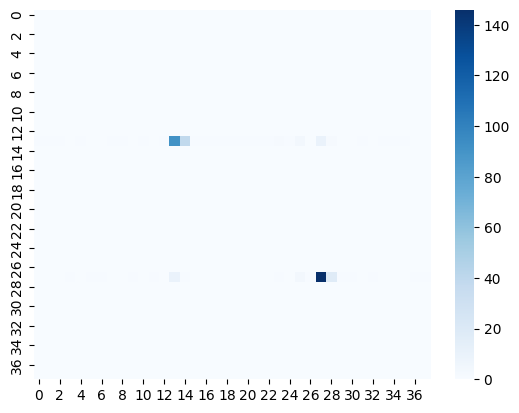}
    \caption{GPT-3.5 responses to AJGT dataset confusion matrix with 5 few-shots. Numbers are removed as most of them, except the highlighted areas, are zeros. Also, responses are removed for clarity purposes}
    \label{fig:gpt35cm_fewshots_5}
\end{figure}

\begin{figure}
    \centering
    \includegraphics[width=.45\textwidth]{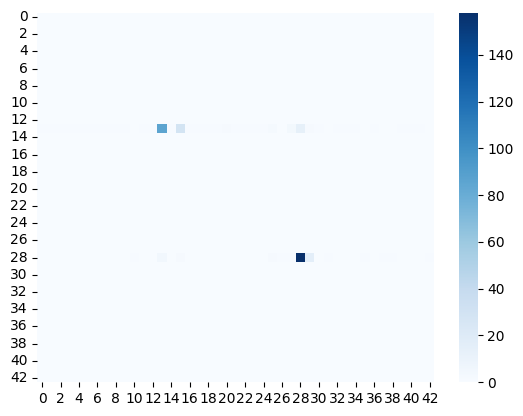}
    \caption{GPT-3.5 responses to AJGT dataset confusion matrix with 10 few-shots. Numbers are removed as most of them, except the highlighted areas, are zeros. Also, responses are removed for clarity purposes}
    \label{fig:gpt35cm_fewshots_10}
\end{figure}

\begin{figure}
    \centering
    \includegraphics[width=.35\textwidth]{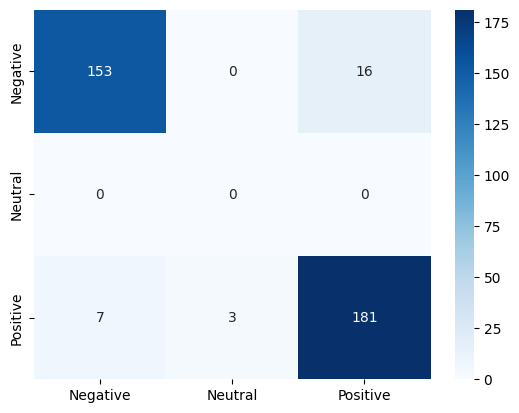}
    \caption{GPT-4 responses to AJGT dataset confusion matrix with 3 few-shots}
    \label{fig:gpt4cm_fewshots_3}
\end{figure}

This section provides extra details and analysis of GPT responses on AGJT dataset to what has been discussed in section \ref{subsec:responses_analysis}. Figures \ref{fig:gpt35cm_fewshots_3}, \ref{fig:gpt35cm_fewshots_5}, and \ref{fig:gpt35cm_fewshots_10} show the confusion matrices of GPT-3.5 responses to AGJT classification task with 3, 5, and 10 few-shots respectively. The activated squares are responses that match the "Negative" and "Positive" responses as instructed by the prompt. From these figures, it can be noticed that as the number of few-shot examples increases, the number of unique responses increases. This is compared to zero-shot settings in Figure \ref{fig:gpt35cm}. The analysis of these responses is already discussed in subsection \ref{subsec:responses_analysis}. On the other hand, Figures \ref{fig:gpt4cm_fewshots_3}, \ref{fig:gpt35cm_fewshots_5}, and \ref{fig:gpt35cm_fewshots_10} show the confusion matrices of GPT-4 responses on the same task. GPT-4 provides a better classification as discussed in the same subsection.

\begin{figure}
    \centering
    \includegraphics[width=.35\textwidth]{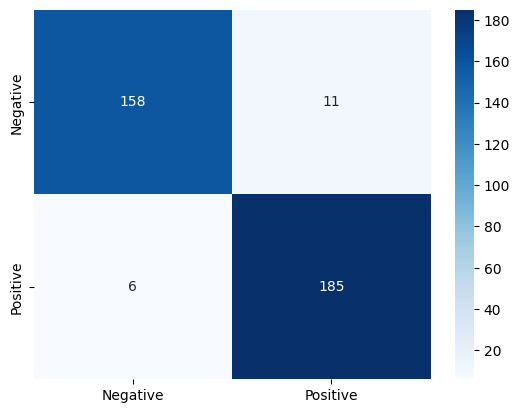}
    \caption{GPT-4 responses to AJGT dataset confusion matrix with 5 few-shots}
    \label{fig:gpt4cm_fewshots_5}
\end{figure}

\begin{figure}
    \centering
    \includegraphics[width=.35\textwidth]{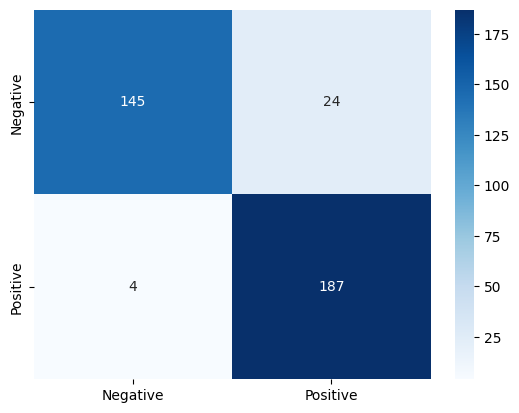}
    \caption{GPT-4 responses to AJGT dataset confusion matrix with 10 few-shots}
    \label{fig:gpt4cm_fewshots_10}
\end{figure}

\section{Breakdown of Diacritization Results}
\label{app:diac-results}

Table \ref{tab:diac-results} shows a detailed breakdown of the results of GPT-4 and GPT-3.5 on the WikiNews diacritization benchmark in comparison to the previously published state-of-the-art results from \cite{mubarak19-highly}.

\input{tables/diacritization}

\section{Examples}
\input{tables/examples}
In Tables \ref{tab:examples_gpt_3} and  \ref{tab:examples_gpt_4}, we show samples for GPT-3.5 and GPT-4 output completion on the test set. Note that, we are truncating the output from the summarization task because it is too long.

\end{document}

%% file: tables/datasets.tex
\begin{table}[!htp]
\centering
\begin{tabular}{lcc}
\toprule
\textbf{Dataset}  & \textbf{Tokens} \\ 
\midrule
 EASC \cite{el2010using}            & 238,993     \\ 
 AJGT \cite{alomari2017arabic}      & 14,465        \\ 
 PADT \cite{smrz2008prague}         & 91,122        \\ 
 APB  \cite{alian2019towards}       & 17,707         \\ 
 UNv1 \cite{ziemski2016united}      & 407,523      \\ 
 BOLT \cite{bies-etal-2014}         & 66,059     \\ 
 WikiNews \cite{darwish17}          & 68,418          \\ 
\bottomrule
\end{tabular}

\caption{Number of input tokens of each dataset. We use the \texttt{tiktoken} library to calculate them.\label{tab:datasets}}

\end{table}

%% file: tables/results.tex
\begin{table*}[!htp]
\centering
\begin{tabular}{llllcc||c}
\toprule

\textbf{Task} & \textbf{Dataset} & \textbf{Test Size} & \textbf{Metric}    & \textbf{GPT-3.5} & \textbf{GPT-4} & \textbf{SoTA}  \\ 
\midrule
\textbf{Summarization} & EASC  & 153     & ($\uparrow$) RougeL  & \underline{\textbf{23.5}}  & 18.25    &  13.3    \\ 
\textbf{Sentiment Analysis}  & AJGT   & 360   & ($\uparrow$) Accuracy    & 86.94  & \underline{90.30}  & \textbf{96.11}    \\ 
\textbf{PoS Tagging}     & PADT   & 680   & ($\uparrow$) Accuracy   & 75.91     & \underline{86.29}     & \textbf{96.83}    \\ 
\textbf{Paraphrasing}    & APB    & 1,010     & ($\uparrow$) BLEU & 4.295     & \underline{6.104}     & \textbf{17.52}  \\ 
\textbf{Translation}     & UNv1   & 4,000     & ($\uparrow$) BLEU  & 35.05     & \underline{38.83}     &  \textbf{53.29} \\ 
\textbf{Transliteration} & BOLT   & 6,653 & ($\uparrow$) BLEU & 13.76     & \underline{27.66} & \textbf{65.88}    \\ 
\textbf{Diacritization} & WikiNews   & 393 & ($\downarrow$) DER  & \underline{10.29} & 11.64 & \textbf{1.21}    \\ 
\bottomrule
\end{tabular}

\caption{Comparing results of GPT-\{3.5, 4\} with SoTA. The test size reflects the number of samples used for evaluating each dataset. The best GPT-based model is \underline{underlined}, and the best result is highlighted in \textbf{bold}.\label{tab:results}}

\end{table*}

%% file: tables/diacritization_finegrained.tex
\begin{table}[!ht]
    \centering
    \begin{tabular}{lccccc}
        \toprule
        \multicolumn{1}{c}{\multirow{2}{*}{\textbf{Domain}}} & \textbf{DER} & \textbf{WER} & \textbf{DER} & \textbf{WER} \\  \cline{2-3} \cline{4-5}
        & \multicolumn{2}{c}{w/ CE} & \multicolumn{2}{c}{w/o CE} \\
        \midrule
        \textbf{Culture} & 9.13 & 30.94  & 8.22 & 22.87 \\ 
        \textbf{Politics} & 9.99 & 31.15  & 9.44  & 24.60 \\ 
        \textbf{Economics} & 10.08 & 33.62  & 9.49  & 26.73 \\ 
        \textbf{Health} & 10.25 & 31.67 & 9.40 & 23.41 \\ 
        \textbf{Sports} & 10.68 & 33.77 & 9.68 & 25.41 \\ 
        \textbf{Science} & 10.70 & 32.95   & 9.71 & 25.03 \\ 
        \textbf{Arts} & 11.55 & 35.64 &  10.08 & 25.59 \\ 
        \hline 
        \textbf{Combined} & 10.29 & 32.74 & 9.39  & 24.77 \\ 
        \bottomrule
    \end{tabular}
    \caption{Fine-grained results of the WikiNews diacritization benchmark, showcasing the performance of \textbf{GPT-3.5} across different domains. The results are presented in ascending order of the Diacritic Error Rate (DER) with case-ending (CE).\label{tab:finegrained-diac-gpt-3.5}}
    
\end{table}

\begin{table}[!ht]
    \centering
    \begin{tabular}{lccccc}
        \toprule
        \multicolumn{1}{c}{\multirow{2}{*}{\textbf{Domain}}} & \textbf{DER} & \textbf{WER} & \textbf{DER} & \textbf{WER} \\  \cline{2-3} \cline{4-5}
        & \multicolumn{2}{c}{w/ CE} & \multicolumn{2}{c}{w/o CE} \\
        \midrule
        \textbf{Culture} & 9.37 & 34.15 & 8.10 & 24.52 \\ 
        \textbf{Health} & 10.74 & 34.80 & 9.10 & 23.82 \\ 
        \textbf{Science} & 11.06 & 37.81 & 9.39 & 27.20 \\ 
        \textbf{Politics} & 11.45 & 36.09 & 10.51 & 27.81 \\ 
        \textbf{Economics} & 11.66 & 39.99 & 10.55 & 30.48 \\ 
        \textbf{Sports} & 14.04 & 42.64 & 12.73 & 32.23 \\ 
        \textbf{Arts} & 14.38 & 42.52 & 12.14 & 30.33 \\ 
        \hline 
        \textbf{Combined} & 11.64 & 38.06 & 10.18 & 27.88 \\ 
        \bottomrule
    \end{tabular}
    \caption{Fine-grained results of the WikiNews diacritization benchmark, showcasing the performance of \textbf{GPT-4} across different domains. The results are presented in ascending order of the Diacritic Error Rate (DER) with case-ending (CE).\label{tab:finegrained-diac-gpt-4}}
    
\end{table}

%% file: tables/prompts.tex
\begin{table}[!htp]
\centering
\begin{tabular}{p{7.5cm}}
\toprule
\textbf{Prompt}  \\
\midrule
Respond only positive or negative sentiment in English  \\\hline
 Predict the sentiment of the following statement in English: choose an option: Positive , Negative \\\hline
Is the sentiment of the following statement Positive or Negative? \\\hline
\RL{
ماهي عاطفة الجملة التالية : أجب} \newline
Positive \RL{أو} Negative \\\hline
You are a helpful assistant that can predict whether a given statement in Arabic is Positive or Negative \\
\bottomrule
\end{tabular}
\caption{The five prompts used for evaluating the AJGT dataset. For each prompt, we vary the text. The fourth prompt is written as a mixture between Arabic and English.}
\label{tab:prompts_eng}
\end{table}

%% file: tables/diacritization.tex
\begin{table*}[ht!]
\centering
\begin{tabular}{lcccc|cccc}
\toprule
\multicolumn{1}{c}{\multirow{3}{*}{\textbf{Model}}} & \textbf{DER} & \textbf{WER} & \textbf{DER} & \textbf{WER}  & \textbf{DER} & \textbf{WER} & \textbf{DER} & \textbf{WER} \\ 
\cline{2-9}
& \multicolumn{2}{c}{w/ Case Ending} & \multicolumn{2}{c|}{w/o Case Ending} & \multicolumn{2}{c}{w/ Case Ending} & \multicolumn{2}{c}{w/o Case Ending} \\ 
\cline{2-9}
& \multicolumn{4}{c|}{Including No Diacritic} & \multicolumn{4}{c}{Excluding No Diacritic} \\
\midrule
\textbf{GPT-3.5}  & 10.29\%   & 32.74\%  & 9.39\% & 24.77\% & 10.38\%   & 26.26\%  & 9.13\% & 18.76\%  \\
\textbf{GPT-4}    & 11.64\%   & 38.06\%  & 10.18\% & 27.88\% & 13.51\%   & 34.35\%  & 12.18\% & 25.82\%  \\ 
\midrule
\textbf{SoTA}     & \textbf{1.21\%}   & \textbf{4.49\%}  & - & \textbf{1.89\%} & - & - & - & -  \\
\bottomrule
\end{tabular}
\caption{Results on the WikiNews diacritization benchmark. DER is diacritic-error-rate and WER is word-error-rate. Results are reported with and without including case-ending as well as including or excluding characters without diacritics during the evaluation. SoTA results from \cite{mubarak19-highly}.\label{tab:diac-results}}

\end{table*}

%% file: tables/examples.tex
\begin{table*}[!htp]
\centering
\begin{tabular}{l|c}
\toprule
\textbf{Task} & \textbf{Example} \\
\hline
\textbf{Diacritization} & \input{examples/gpt3_5/diacritization} \\ \hline
\textbf{Sentiment Analysis} & \input{examples/gpt3_5/sentiment_analysis} \\ \hline
\textbf{Paraphrasing} & \input{examples/gpt3_5/paraphrasing} \\ \hline
\textbf{PoS Tagging} & \input{examples/gpt3_5/pos_tagging}\\ \hline
\textbf{Translation} & \input{examples/gpt3_5/translation}\\ \hline
\textbf{Summarization} & \input{examples/gpt3_5/summarization}\\ \hline
\textbf{Transliteration} & \input{examples/gpt3_5/transliteration}\\ \hline

\bottomrule
\end{tabular}
\caption{Examples of Input and Output pairs for GPT-3.5 across the 7 Arabic NLP Tasks explored in this paper. The input text for each task is displayed above the divider, while the corresponding output generated by GPT-3.5 is shown below the divider.\label{tab:examples_gpt_3}}

\end{table*}

\begin{table*}[!htp]
\centering
\begin{tabular}{l|c}
\toprule
\textbf{Task} & \textbf{Example} \\
\hline
\textbf{Diacritization} & \input{examples/gpt_4/diacritization} \\ \hline
\textbf{Sentiment Analysis} & \input{examples/gpt_4/sentiment_analysis} \\ \hline
\textbf{Paraphrasing} & \input{examples/gpt_4/paraphrasing} \\ \hline
\textbf{PoS Tagging} & \input{examples/gpt_4/pos_tagging}\\ \hline
\textbf{Translation} & \input{examples/gpt_4/translation}\\ \hline
\textbf{Summarization} & \input{examples/gpt_4/summarization}\\ \hline
\textbf{Transliteration} & \input{examples/gpt_4/transliteration}\\ \hline

\bottomrule
\end{tabular}
\caption{Examples of Input and Output pairs for GPT-4 across the 7 Arabic NLP Tasks explored in this paper. The input text for each task is displayed above the divider, while the corresponding output generated by GPT-4 is shown below the divider.\label{tab:examples_gpt_4}}

\end{table*}

%% file: examples/gpt3_5/diacritization.tex
\makecell{
\RL{فيلم جاذبية يتصدر ترشيحات جوائز الأكاديمية البريطانية لفنون الفيلم والتلفزيون}\\
\hline
\RL{فيلم جَاذِبِيَّة يَتَصَدَّرُ تَرْشِيحَاتِ جُوَائِزِ الْأَكَادِيمِيَّةِ الْبَرِيْطَانِيَّةِ لِفَنُوْنِ الْفِيْلْمِ وَالتِّلْفِزْيُونِ}
}

%% file: examples/gpt3_5/sentiment_analysis.tex
\makecell{
\RL{فما  اجمل  هذا  الدين} \\
\hline
Positive
}

%% file: examples/gpt3_5/paraphrasing.tex
\makecell{
\RL{سافر أبي إلى تونس البارحة.} \\
\hline
\RL{سافر والدي إلى تونس يوم أمس.}
}

%% file: examples/gpt3_5/pos_tagging.tex
\makecell{
\RL{سوريا : تعديل وزاري واسع يشمل 8 حقائب} \\
\hline
\RL{سوريا}:NOUN,
: :PUNCT,
\RL{تعديل}:NOUN,
\RL{وزاري}:ADJ,
\\
\RL{واسع}:ADJ,
\RL{يشمل}:VERB,
8:NUM,
\RL{حقائب}:NOUN,
}

%% file: examples/gpt3_5/translation.tex
\makecell{
\RL{الجلسة 7439 المعقودة في 11 أيار/مايو ٢٠١٥.} \\
\hline
Session 7439 held on May 11, 2015.
}

%% file: examples/gpt3_5/summarization.tex
\makecell{
\RL{﻿لودفيج فان بيتهوفن مؤلف موسيقي ألماني ولد عام 1770 ... } \\
\hline
\RL{لودفيج فان بيتهوفن هو مؤلف موسيقي ألماني ولد في بون ... }
}

%% file: examples/gpt3_5/transliteration.tex
\makecell{
Al facebook bat3k a ? \\
\hline
\RL{الفيسبوك بتاعك إيه؟}
}

%% file: examples/gpt_4/diacritization.tex
\makecell{
\RL{فيلم جاذبية يتصدر ترشيحات جوائز الأكاديمية البريطانية لفنون الفيلم والتلفزيون}\\
\hline
\RL{فِيلَم جَاذِبِيَّة يَتَصَدَّر تَرْشِيحَات جَوَائِز الأَكَادِيمِيَّة البَرِيطَانِيَّة لِفُنُونِ الفِيلَمِ وَالتِلْفِزْيُون}
}

%% file: examples/gpt_4/sentiment_analysis.tex
\makecell{
\RL{فما  اجمل  هذا  الدين} \\
\hline
positive
}

%% file: examples/gpt_4/paraphrasing.tex
\makecell{
\RL{سافر أبي إلى تونس البارحة.} \\
\hline
\RL{توجه والدي إلى تونس يوم أمس.}
}

%% file: examples/gpt_4/pos_tagging.tex
\makecell{
\RL{سوريا : تعديل وزاري واسع يشمل 8 حقائب} \\
\hline
\RL{سوريا}:NOUN,
: : ,
\RL{تعديل}:NOUN,
\RL{وزاري}:ADJ,
\\
\RL{واسع}:ADJ,
\RL{يشمل}:VERB,
8:NUM,
\RL{حقائب}:NOUN,
}

%% file: examples/gpt_4/translation.tex
\makecell{
\RL{الجلسة 7439 المعقودة في 11 أيار/مايو ٢٠١٥.} \\
\hline
Session 7439 held on May 11, 2015.
}

%% file: examples/gpt_4/summarization.tex
\makecell{
\RL{﻿لودفيج فان بيتهوفن مؤلف موسيقي ألماني ولد عام 1770 ... } \\
\hline
\RL{لودفيج فان بيتهوفن موسيقي ألماني عبقري ولد في 1770 ... }
}

%% file: examples/gpt_4/transliteration.tex
\makecell{
Al facebook bat3k a ? \\
\hline
\RL{على الفيسبوك بتعك ؟}
}